# Optimized Trajectory Planning for USVs Under Ocean Currents

Behzad Akbari, *Member, IEEE*, Ya-Jun Pan, *Senior Member, IEEE,* Shiwei Liu, and Tianye Wang

*Abstract —* **The proposed work focuses on the trajectory planning for Unmanned Surface Vehicles (USVs) in the ocean environment, considering various spatiotemporal factors such as ocean currents and other energy consumption factors. The paper uses Gaussian Process Motion Planning (GPMP2), a Bayesian optimization method that has shown promising results in continuous and nonlinear motion planning algorithms. The proposed work improves GPMP2 by incorporating a new spatiotemporal factor for tracking and predicting ocean currents using a spatiotemporal Bayesian inference. The algorithm is applied to the USV path planning and is shown to optimize for smoothness, obstacle avoidance, and ocean currents in a challenging environment. The work is relevant for practical applications in ocean scenarios where optimal path planning for USVs is essential for minimizing costs and optimizing performance.**

## I. Introduction

An Unmanned Surface Vehicle (USV) is a self-operating watercraft with significant roles in various marine applications, including military operations, surveillance, and search and rescue missions. Trajectory planning is a crucial responsibility for autonomous USVs, ensuring mission success while minimizing energy consumption. Effective marine navigation for a USV involves a two-fold approach: a global planner for initial trajectory establishment and a local planner for real-time event management.

The global planner's role primarily revolves around generating an initial trajectory based on prior information, including static geographic features and meteorological data. This initial trajectory, computed centrally, serves as the foundation for mission commencement. However, as the USV interacts with its dynamic environment, a local planner becomes essential. The local planner addresses uncertainties and updates the initial trajectory based on real-time events and constraints. Key considerations for the local planner encompass dynamic factors like collision avoidance and energy consumption, which can change over time.

Despite the increasing interest in USV path planning, significant challenges remain, including considering realistic environmental impacts such as winds and surface currents. Furthermore, only a minority of algorithms consider these environmental characteristics and fail to consider their spatiotemporal properties. Several algorithms have recently been developed to address these challenges, including Gaussian Process Motion Planning (GPMP2), which can generate short and smooth paths in a large-scale or high-dimensional space. GPMP2 is non-parametric, computationally efficient, and compatible with Markov models [1] [2]. While GPMP2's capabilities make it a promising solution for trajectory planning in various autonomous vehicles, including USVs, there is still a need for improvements to tackle complex environments like the ocean.

This study introduces an energy consumption factor specific to ocean currents within the GPMP2 algorithm. A spatiotemporal Gaussian Process model is utilized to represent the ocean current. Incorporating a spatiotemporal filter along with the energy consumption factor produces a seamless, ongoing, and adaptable trajectory for each step, reducing the probability of collisions and energy consumption. The system's interaction and stability are enhanced by utilizing a local planner to optimize the entire trajectory in each time step. The factor graph model of the trajctory allows us to address intricate optimization problems like GPMP2 in real-time. This paper makes several significant contributions, including:

1) Introducing a novel spatiotemporal Gaussian process inference technique to track ocean currents throughout the trajectory in each time step.
2) Utilizing the estimated ocean current vector in combination with GPMP2 to generate a trajectory that minimizes energy consumption and ensures collision avoidance.
3) The ocean current dataset for Halifax Harbor is utilized for training the Gaussian Process (GP) model and determining the kernel parameters and transition matrices.

## II. Related Works

When it comes to planning paths for USVs, commonly employed approaches include optimization, heuristic searching, or a blend of both. One of the optimization methods widely used is the Artificial Potential Field (APF), which generates a potential field of the environment and constraints to create a path. The destination generates global forces, while obstacles and constraints generate local repulsive forces. A multi-layered APF model that minimizes energy consumption is introduced in [3] for USV path planning. In dynamic environments, the collision risk is measured by local repulsive forces based on the distance between the USV and the closest obstacles. Wind and currents, which are external and time-varying factors, can also create collision risks. The Fast Marching Method (FMM) has been proposed in [4] [5] as a solution to generate the shortest path while considering these factors. Additionally, a variation of FMM called anisotropic fast marching (AFM) has been combined with GPMP2 and augmented with factors for wind or ocean currents [6]. The AFM method expands upon the capabilities of the FMM by accommodating

B. Akbari and Y.J. Pan are with the Department of Mechanical Engineering, Dalhousie University, Halifax, NS, Canada, B3H 4R2. (Email: bh888652@dal.ca, yajun.pan@dal.ca). Shiwei Liu and Tianye Wang are with Marine Thinking, 1096 Marginal Road, Halifax, NS B3H 4N4. (Email: shiwei.liu@marinethinking.com, tianye.wang@marinethinking.com).

anisotropic media, wherein the speed of the wavefront can vary across different directions. This extension proves particularly advantageous in path planning where the medium involves direction-dependent factors, such as external forces impacting the movement. Common heuristic methods for USV path planning include well-known algorithms such as Ant Colony Optimization (ACO), A-star (A*), and Dijkstra's algorithm [7] [8]. Additionally, sampling-based planning methods provide an alternative to graph-based global planners, eliminating the need for initializing the state space with a grid map before planning begins [9]. There are two main categories of sampling-based algorithms:

1. Multi-query planners, exemplified by the Probabilistic Road Map (PRM) [10].
2. Single-query planners, like the Rapidly exploring Random Tree (RRT) [11].

The RRT algorithm incrementally constructs a tree by populating the entire search space with randomly sampled points. This makes it particularly effective for finding solutions in high-dimensional spaces. However, it may produce suboptimal paths, and repeating the search process under the same conditions can yield significantly different final paths. To address these limitations, the RRT* algorithm was introduced. RRT* incrementally reconnects the tree around a new node and its neighboring nodes, resulting in improved solution quality [12]. In our specific application, where we have prior knowledge of start and goal points, and computational resources are efficient [13], we chose to employ the RRT* algorithm to obtain the initial path.

**Table I:** Abbreviations adopted in this manuscript.

| Abbreviation | Description |
|---|---|
| USV | Unmanned Surface Vehicle |
| GP | Gaussian Process [14] |
| GPMP | Gaussian Process Motion Planning [2] [15] |
| MRPP | Multi-Robot Path Planning |
| APF | Artificial Potential Field |
| FMM | Fast Marching Method |
| ACO | Ant Colony Optimisation |
| MAP | Maximum A Posteriori |
| LTV-SDE | Linear Time-varying Stochastic Differential Equation |
| PRM | Probabilistic Road Map |
| RRT | Rapidly-exploring Random Tree [11] |
| BFGS | Broyden-Fletcher-Goldfarb-Shanno |
| AFM | Anisotropic Fast Marching |
| GPGN | Gaussian process Gauss-Newton |

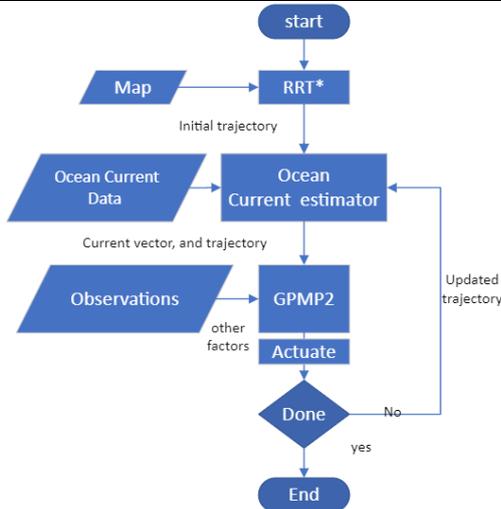

**Fig. 1.** The core iterative cycle of the algorithm

## III. PROBLEM FORMULATION

The optimization of USV trajectories involves building upon the initial path created by the RRT* algorithm. To tackle this challenge of ocean currents, we break the problem down into two essential components. First, we employ Gaussian Process inference, akin to the GPMP2 approach, to fine-tune the trajectory. Second, we utilize ocean current tracking to estimate the ocean current vector along the trajectory. These two components collaborate to derive a trajectory that minimizes the impact of ocean currents on energy consumption, the framework of the algorithm explained in Fig.1.

### A. USV Trajectory Optimization

The modeling method employed for USV trajectory optimization is a variable-based model. In this type of model, estimation values are defined using random variables [16]. The goal of inference is to find the best estimation that satisfies all constraints and maximizes the posterior density of those variables. In USV path planning each point along the trajectory is represented by a set of random variables, encompassing both position and velocity. Leveraging the smooth nature of ocean phenomena, we employ Gaussian process to model the USV trajectory. A GP can be considered a generalization of a Gaussian random variable in a continuous-time domain [17]. GP state functions $x(t)$, is described by a mean function, $\boldsymbol{\mu}(t)$, and a kernel function, $\boldsymbol{K}(t,t')$. The system model is described as follows:

$$\mathbf{x}(t) \sim GP(\boldsymbol{\mu}(t), \mathbf{K}(t,t')), \quad (1)$$

#### 1) GP inference

Evaluating the continues GP at discrete instants of time results in a joint Gaussian random variable. Usually, a GP function can be represented by a vector value function with length of $n_s$ and their interpolations. An efficient gradient-based optimization algorithm can be used to compute the GP interpolation. Formally, trajectory optimization aims to determine the optimal trajectory from all feasible trajectories while satisfying all constraints and minimizing the event's cost. GP regression is performed by a maximum posteriori (MAP) inference, where the most likely trajectory is the mean of the posterior distribution conditioned on the events and corresponding measurements. The MAP estimate of the trajectory can be computed through the Gaussian process Gauss-Newton (GPGN) [18]. To define the objective function for discrete (valued function) trajectory $\mathbf{x}$, we assume that the size of the state is $n_s$ and there are $N_f$ factors(constraints).

$$\mathbf{x} = \begin{bmatrix} x(1), \dot{x}(1), \\ \vdots \\ x(n_s), \dot{x}(n_s) \end{bmatrix}, \boldsymbol{\mu} = \begin{bmatrix} \mu(1) \\ \vdots \\ \mu(n_s) \end{bmatrix}, \mathbf{K}_{n_s * n_s} \quad (2)$$

A posterior function for discrete state $\mathbf{x}$, given all measurements regarding the $N_f$ factors $\mathbf{z} = \{z_1, z_2, \ldots z_{N_f}\}$ can be expressed as $P(\mathbf{x}|\mathbf{z})$. The optimal trajectory coming from the MAP estimation of the posterior function as:

$$\mathbf{x}^* = \underset{\mathbf{x}}{\mathrm{argmax}} \ P(\mathbf{x}|\mathbf{z}). \quad (3)$$

The proportion of posterior distribution of **x** given **z** can be derived from the prior P(**x**) and likelihood L(**z**|**x**) by Bayes rule as:
$$P(\mathbf{x}|\mathbf{z}) \propto P(\mathbf{x})L(\mathbf{z}|\mathbf{x}). \quad (4)$$
This formula will be represented as the product of a series of independent factors [19],
$$P(\mathbf{x}|\mathbf{z}) \propto P(\mathbf{x})L(z_1|\mathbf{x})\ldots L(z_{N_f}|\mathbf{x}) \propto \prod_{n_f=1}^{N_f} f_{n_f}(\mathbf{x}). \quad (5)$$

In cases where we need to represent each event as a factor $f_{n_f}(\mathbf{x})$ on state subsets **x**, such as spatiotemporal factors like ocean currents, we often work with estimations due to the unavailability of actual values. To efficiently address the maximum a posteriori (MAP) problem, which encompasses all factors, we can leverage the inherent sparsity within the factor graph. This is demonstrated in [19]

*2) Interpolating and GP Markov model*

For interpolating, based on [19] [20], a linear time-varying stochastic differential equation form of a GP can be written as:
$$\dot{\mathbf{x}}(t) = \mathbf{A}(t)\mathbf{x}(t) + \boldsymbol{\vartheta}(t) + \mathbf{F}(t)\mathbf{w}(t), \quad (6)$$

where $\mathbf{x}(t)$ represents the state in continues domain, $\boldsymbol{\vartheta}(t)$ is the known system control input, $\mathbf{A}(t)$ and $\mathbf{F}(t)$ are time-varying transition functions of the system, and $\mathbf{w}(t)$ is generated by a white noise process. The white noise process is itself a zero-mean GP:
$$\mathbf{w}(t) \sim GP(0, \boldsymbol{Q}_C \delta(t-t')), \quad (7)$$
where $\boldsymbol{Q}_C$ is the power-spectral density matrix and $\delta(t-t')$ is the Dirac delta function. The solution to the initial value problem of this Linear Time-Varying Stochastics Differential Equation (LTV-SDE) is in the form of mean and covariance [20]:
$$\widetilde{\boldsymbol{\mu}}(t) = \boldsymbol{\Phi}(t, t_0)\boldsymbol{\mu}_0 + \int_{t_0}^{t} \boldsymbol{\Phi}(t, s)\boldsymbol{\vartheta}(s)ds, \quad (8)$$

$$\widetilde{\mathbf{K}}(t, t') \\ = \boldsymbol{\Phi}(t, t_0)\mathbf{K}_0\boldsymbol{\Phi}(t', t_0)^T \\ + \int_{t_0}^{\min(t,t')} \boldsymbol{\Phi}(t, s)\mathbf{F}(s)\boldsymbol{Q}_C\mathbf{F}(s)^T\boldsymbol{\Phi}(t', s)^T ds, \quad (9)$$

where $\boldsymbol{\Phi}$ is the state transition matrix, and $\boldsymbol{\mu}_0$ and $\mathbf{K}_0$ are the mean and covariance, at $t_0$ respectively. The Markov property of Eq. (8) results in the sparsity of the inverse kernel matrix $\mathbf{K}^{-1}$, which allows for a fast inference.

*3) MAP inference*

Considering the valued function **x** corresponding to continues function $\mathbf{x}(t)$, the GP prior represents the dynamics of a system, and its proportion can be written as follows:
$$P(\mathbf{x}) \propto \exp\left\{-\frac{1}{2}\|\mathbf{x}-\boldsymbol{\mu}\|_{\mathbf{K}}^2\right\}. \quad (10)$$

Also the likelihood functions in Eq. (5) formulates constraints as events that the GP function has to obey. For example, the likelihood function of obstacle avoidance in a path planning scenario indicates the probability of being free from obstacles. All the likelihood functions need to be limited to the exponential family, given by:
$$L(\mathbf{x}; z_{n_f}) \propto \exp\left\{-\frac{1}{2}\|h_{n_f}(\mathbf{x})\|_{\boldsymbol{\Sigma}_{n_f}}^2\right\}, \quad (11)$$

where $h_{n_f}(\mathbf{x})$ is a vector-valued hinge cost for an event $n_f$ and $\|.\|_{\boldsymbol{\Sigma}_{n_f}}^2$ is the Mahalanobis distance with hyper parameters (covariance) $\boldsymbol{\Sigma}_{n_f}$. In this paper, we introduced a new ocean current factor. For other essential factors required for trajectory planning, such as prior factors, obstacle avoidance, and kenematic limitations, we based our approach on the paper in [1] [19] [21].

Using Eqs. (5), (10), and (11), after operating $-\log(.)$ and converting $argmax$ to $argmin$, the MAP posterior estimation can be formulated as
$$\mathbf{x}^* = \underset{\mathbf{x}}{\operatorname{argmin}}\Big\{\frac{1}{2}\|\mathbf{x}-\boldsymbol{\mu}\|_{\mathbf{K}}^2 + \frac{1}{2} \\ \|h_2(\mathbf{x})\|_{\boldsymbol{\Sigma}_2}^2 + \cdots + \\ \|h_{n_f}(\mathbf{x})\|_{\boldsymbol{\Sigma}_{n_f}}^2\Big\}, \quad (12)$$

which is a nonlinear least square problem that can be solved by iterative algorithms such as Gauss-Newton or Levenberg-Marquardt until convergence. The corresponding linear equation is well-known as follows:
$$\mathbf{x}^* = \underset{\mathbf{x}}{\operatorname{argmin}} \|A x - \boldsymbol{b}\|^2, \quad (13)$$
where $A \in \mathbb{R}^{n_f \times n_s}$ is the measurement Jacobian consisting of $n_f$ measurement rows and $\boldsymbol{b}$ is an $n_s$-dimensional vector computable like iSAM2 [21]. The factor graph representing the trajectory optimization is illustrated in Fig. 2.

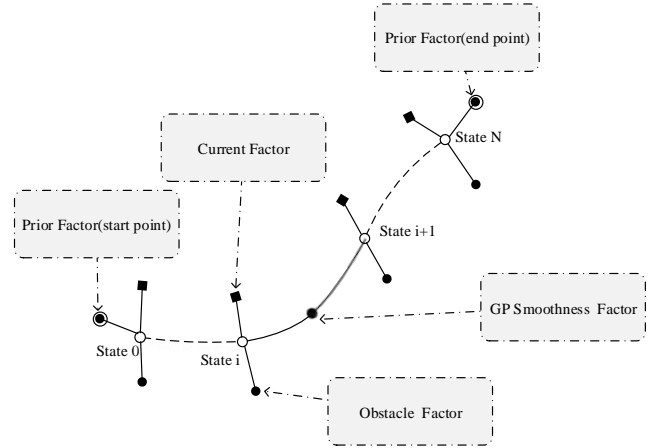

**Fig. 2.** A factor graph of a trajectory optimization problem for USV path planning. Note that the support states are marked as white circles, and there are four main types of factors, namely prior factors on the start and goal states, GP factors, Ocean Current factors, obstacle factors, and interpolation between states.

*B. Tracking The Energy Consumption*

Energy consumption can be determined by calculating the force required to maintain a specified speed and heading under varying environmental conditions. Regarding the tracking of energy consumption related to ocean currents, we discussed the factors associated with ocean currents and the Gaussian model representing them.

*1) Ocean Current Factor*

Ocean Current factor depends on multiple uncertain parameters, including the speed and direction of the current, as well as other pertinent environmental and operational factors. Within an oceanic setting, these factors encompass water density, ocean currents, waves, and wind, all of which can exert an influence on the motion of a USV. Importantly, these factors exhibit spatiotemporal variability, meaning they depend on both location and time. To illustrate, data from Halifax Harbor reveals significant fluctuations in ocean currents over a 24-hour period due to tidal changes. The impact of these forces on a USV is profoundly affected by the angles at which they act upon the vehicle, a characteristic largely determined by the USV's shape. Long and pointed Unmanned Surface Vehicles (USVs) may be less affected by ocean currents because they have a reduced surface area exposed to these forces. The likelihood of energy consumption is directly linked to the deviation between the desired position and the actual position, as illustrated in the Hinge Loss function $h_c(\mathbf{x})$, as:

$$L(\mathbf{x}; z_c) \propto \exp\left\{-\frac{1}{2} \parallel h_c(\mathbf{x}) \parallel_{\Sigma_c}^2\right\}, \quad (14)$$

Where for the indexed state $x_i$ in trajectory $\mathbf{x}$:

$$h_c(x_i) = \begin{cases} -d(x_i) + \epsilon & if \parallel d(x_i) \parallel > \epsilon \\ 0 & if \parallel d(x_i) \parallel \leq \epsilon \end{cases}, \quad (15)$$

$d(x_i)$ is deviation value between the desired position and the actual position in two dimensions, and $\epsilon$ is the maximum allowed deviation in each axes. The function ensures that the loss is only incurred when the norm of deviation exceeds the threshold. If the magnitude of deviation exceeds the threshold ($\parallel d(x_i) \parallel > \epsilon$), the loss is proportional to the amount by which it exceeds the threshold. The hinge loss function penalizes the USV's trajectory when it deviates too far from the desired path due to the ocean current. This loss function can be used as part of the GP function optimization to guide the USV toward its desired path while considering the influence of ocean currents. In our proposed paper, we calculate $d(x_i)$, between a state $x_i$ and the subsequent state $x_{i+1}$. This calculation is based on the discrepancy between the actual velocity and the ocean current velocity, adjusted by a factor represented by $C(\alpha)$, as illustrated in Fig. 3. The computation of $d(x_i)$ vector is expressed as follows:

$$d(x_i) = (v_a(x_i) - v_c(x_i))C(\alpha) \; ; \; C(\alpha) \propto |sin(\alpha)|. \quad (16)$$

In this context, $v_a(x_i)$ signifies the actual velocity, and $v_c(x_i)$ represents the ocean current velocity at both axes for indexed state $x_i$. The factor $C(\alpha)$ corresponds to the magnitude proportion of the ocean current's force exerted on the USV's body, and it is directly linked to the angle $\alpha$. In this research, a magnitude of a sine function $|sin(\alpha)|$ is utilized to compute $C(\alpha)$. This choice is made because it accurately captures how the impact of the ocean current on the USV's body varies, increasing as the angle approaches a right angle and diminishing when it deviates from that point.

Accurately calculating the energy consumption factor for the entire trajectory requires incorporating current velocities at different positions along the path. However, these velocities are not always available for all locations and times. Since deviations in ocean currents often follow a Gaussian distribution, we can potentially estimate them accurately using Bayesian filtering. In this section, we introduce an extended target tracking method to estimate spatiotemporal ocean currents along the trajectory. This approach allows us to track the energy consumption factor continuously in both space and time, drawing inspiration from the methods described in references [22] and [23].

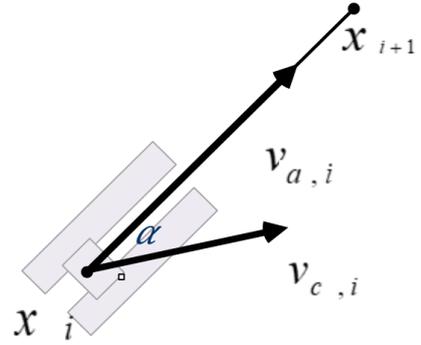

**Fig. 3.** The configuration of the USV and the impact of the current on its motion are described by the actual velocities $v_{a,i}$ and the current velocity $v_{c,i}$. These velocities refer to the two next indexed states $x_i$ and $x_{i+1}$ along the trajectory.

Fig. 4 illustrates the concept of estimating the energy consumption factor by tracking the ocean current function. The estimation, using uncertain measurements (Fig. 4.A) and the given previous trajectory (Fig. 4.B), helps us find the continuous function representing the ocean current values (Fig. 4.C).

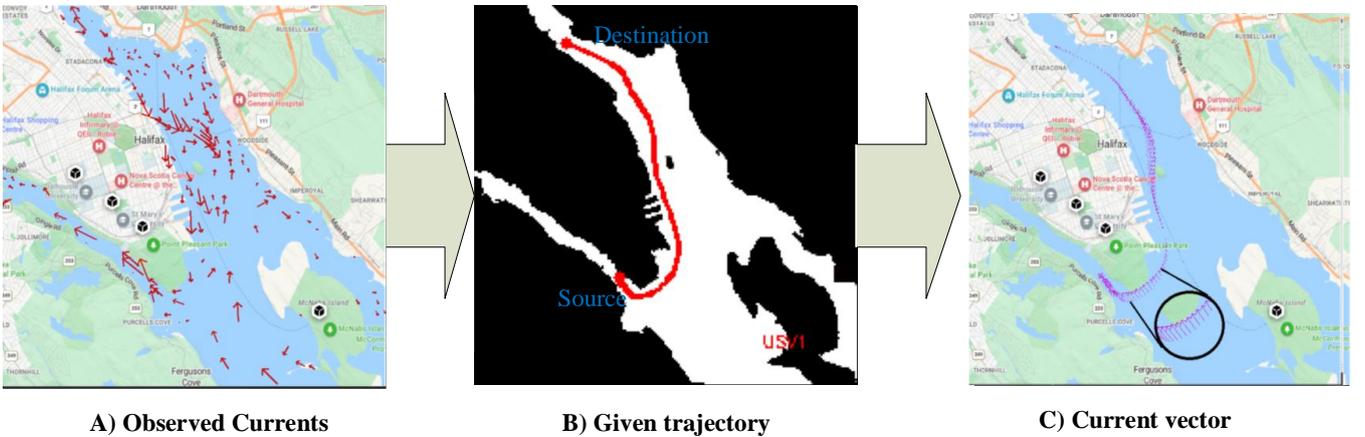

**A) Observed Currents**      **B) Given trajectory**      **C) Current vector**

**Fig. 4.** Using the observed measurements in subfigure A) and the previous given trajectory B), pseudo measurements are estimated for the Currents in subfigure C). The outcome of subfigure C) will be utilized to calculate the trajectory cost and find the updated optimal path.

*2) Ocean Current Gaussian Model*

For a given index $x_i$, the ocean current $v_c(x_i)$ can be represented as a vector. This vector indicates the ocean current velocity in the both axes at each point along the trajectory. Notice that in this section $v_c(x_i)$ is the state that we are trying to estimate for the index $x_i$. The Gaussian nature of ocean currents allows us to employ kernel-based spatiotemporal Gaussian process inference similar to [22] to track the Current's vectors in both space and time. The model includes a limited number of index points $x_i$, which corresponds to the trajectory locatins, and the latent function $v_c(x_i)$ represents the ocean currents at those points. In our Gaussian process model, for simplicity, we design the dynamics of the ocean current vector as a separable kernel.

$$\kappa(x, x'; t, t') = \kappa_x(x, x')\kappa_t(t, t'), \quad (17)$$

where $\kappa(x, x'; t, t')$ is the Spatio-temporal covariance kernel, $\kappa_x(x, x')$ the spatial kernel, and $\kappa_t(t, t')$ represents the temporal covariance kernel. In this paper, the spatial kernel for two individual indexes $(x, x')$ is considered a Radial Basis Function (RBF) kernel as [24]:

$$\kappa(x, x')_x = \sigma^2 \exp(-\frac{1}{2}\frac{(x-x\prime)^2}{\ell^2}). \quad (18)$$

Hyperparameters of the kernel $\sigma$, $\ell$ are learned from limited number of the training data. For the temporal covariance kernel, we primarily use the Whittle-Matern kernel [24] will explain in Appendix A, Eq. (25). The proposed model is converted to an equivalent state-space representation. The evolution of states is modeled as follows:

$$\mathbf{v}_k = \mathbf{F}_k \mathbf{v}_{k-1} + \mathbf{L}_k \mathbf{w}_k, \quad (19)$$
$$\mathbf{w}_k \sim \mathcal{N}(0, \mathbf{Q}_k(\mathbf{x}, \mathbf{x}'; T_s)),$$
$$\mathbf{v}_k = [v_c(x_1), v_c(x_2), \dots, v_c(x_{n_f})]^\top \quad (20)$$

where $\mathbf{v}_k$ is the state of the Ocean Current in which each corresponds to $n_f$ distinct index points same as the trajectory. The state transition matrix is denoted by $\mathbf{F}_k$, and $\mathbf{w}_k$ is a zero-mean white Gaussian noise with covariance $\mathbf{Q}_k$. Note that all coordinates are in Cartesian space with a single global origin. To find the hyper parameters we train the model similar to a regular GP, minimizing the log marginal likelihood of measurements [25]. The training results include the kernel parameters and transition matrices.

*3) Measurement Model for Ocean Currents*

Ocean currents can be observed using a limited number of environmnetal sensors around the area. Usually, the cardinality and position of the measurements are both unknown and time-varying. In addition, the measurement origin uncertainty should be considered. Considering a set of $n_f$ distinct indexes, we want to estimate the Current states, $\mathbf{v}_k$ in time step k corresponding to these noisy measurements $\mathbf{z}_k$. The pseudo measurements corresponding to index points $\mathbf{x}_k$ (as in Fig.4.C)) is given based on Eq.(21):

$$\mathbf{z}_k = \mathbf{H}_k \mathbf{v}_k + \upsilon_k, \quad \upsilon_k \sim \mathcal{N}(0, \mathbf{R}_k), \quad (21)$$

where $\mathbf{H}_k$ is a measurement transition matrix and $\upsilon_k$ is the zero-mean Gaussian measurement noise.

The measurement likelihood is given for the measurements according to the joint distribution probability density of GP as Eqs.(23) and (24).

$$p(\mathbf{z}_k|\mathbf{v}_k) \sim \mathcal{N}(\mathbf{z}_k; \mathbf{H}_k\mathbf{v}_k, \mathbf{R}_k). \quad (22)$$

The implementation of the measurement transition matrix $\mathbf{H}_k$ and measurement covariance $\mathbf{R}_k$ is derived from:

$$\mathbf{H}_k = (K(\mathbf{x}_k, \mathbf{x}^f))[K(\mathbf{x}^f, \mathbf{x}^f)]^{-1}, \quad (23)$$

$$\mathbf{R}_k = \left(K(\mathbf{x}_k, \mathbf{x}_k)\right) + R - \quad (24)$$
$$\left(K(\mathbf{x}_k, \mathbf{x}^f)\right)[K(\mathbf{x}^f, \mathbf{x}^f)]^{-1}\left(K(\mathbf{x}^f, \mathbf{x}_k)\right).$$

Notice that $\mathbf{x}^f$ are the indexes that we are trying to estimate and $\mathbf{x}_k$ are indexes that we observed and $R$ is a constant noise.

Based on the Current model and the measurement model discussed above, we can predict unobserved values of the Current $\mathbf{v}_k$ at index points $\mathbf{x}^f$ using observed measurements Fig. 5. For additional details regarding the tracking of ocean currents, please refer to Appendix A.

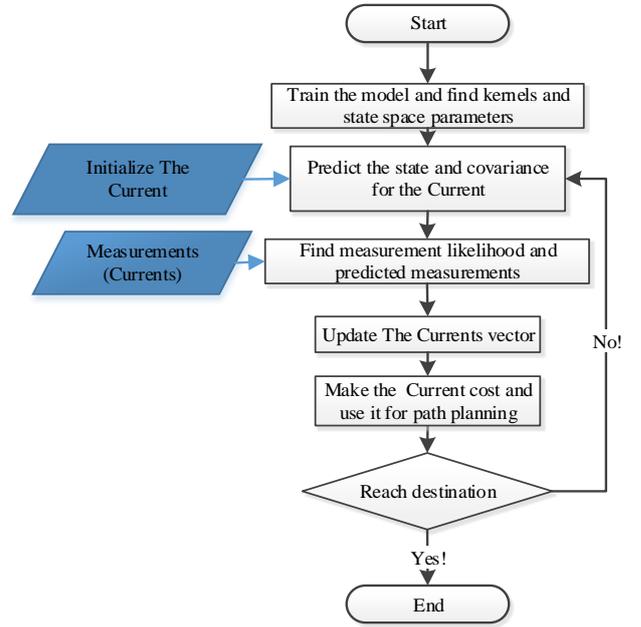

**Fig. 5.** Tracking the ocean current in the USV path planning

IV. SIMULATION SCENARIO

*A. Training the Model*

We utilized a data set of the Halifax harbor ocean data from [26] [27]. Halifax Harbour is an estuary that is partly enclosed and heavily influenced by tidal patterns, wind activity, and various water discharges from sources such as the Sackville River, sewage systems, and coastal water distribution. The strongest Current in the harbor can reach speeds between 15-35 cm/s in the Narrows area and 5-15 cm/s along the boundary between the Inner and Outer Harbour. We employed distinct time and space kernels. Training was carried out using the Gpy Gaussian Process Python library and the Broyden-Fletcher-Goldfarb-Shanno (BFGS) optimization algorithm [28]. The outcome of the training process is a set of hyperparameters that can be applied to calculate the state space transition matrices. In the spatial domain, we trained and optimized the RBF kernel, similar to Eq. (18), and in the temporal domain, the Matern kernel, equivalent to Eq. (25), using the training data $\{\mathbf{x}_d^*, \mathbf{z}_d^*\}_{d=1}^D$.

The outcome of the training algorithm, Algorithm 2 is a set of transition variables in the time domain and an optimized kernel in the space domain. The optimized kernel, represented by $K(.)$, will be utilized to determine measurement transition matrix and likelihood matrix in Eqs. (23) and (24).

*B. USV path planning scenario*

Let's consider a USV deployed for marine research and monitoring purposes in the Halifax area. Maps are created in the area with two longitudes and latitudes representing an area (-63.6177, 44.6745, -63.5037, 44.5978) as in Fig. 4. The data used for the simulation and training of ocean currents are gathered from the sources referenced as [26] and [27]. The dataset comprises the hourly measurements of current velocity throughout a specific day. The goal is to determine the most optimal route for the USV to travel from its starting point to the destination. In this particular scenario, the USV is required to navigate through a complex maritime environment, which may consist of obstacles like islands, reefs, or shipping lanes. To assess the algorithm's performance, we conducted tests in two distinct ocean current scenarios, A and B. The key difference between A and B is the reversal of the current in a specific section, as indicated by a circle in Fig. 6. The paths resulting from these tests are depicted in Fig. 6, illustrating the outcomes under Current settings A and B, respectively, in two ways. First, by adapting the result path to the changing currents, we demonstrate how to avoid adverse current conditions. Secondly, the comparison underscores the differences in consumption factor improvement between paths initially formulated without considering ocean currents (represented by a dotted line) and those modified to account for ocean currents in subsequent steps (represented by a solid line). In both current scenarios A and B, we observed improvements in the consumption factor of approximately 7% and 16%, respectively, when the USV avoided paths in areas with strong currents.

The consumption factor was calculated using the methodology described in Eq. 16. This approach also accounts for the signed effect of the current on the dynamics of the USV. In addition to considering ocean currents, the path planning algorithm aims to ensure collision avoidance and optimize smoothness, like a standard GPMP method.

The path creation process begins by establishing a global path without considering ocean currents (dotted style line) using relevant data sources, such as available maps. After the initial setup, we update the trajectory at each step to reach the destination considering ocean currents A and B. In our simulations, we assumed the location of the USV was known, but in real-world applications, a GPS sensor could be used for localization. Based on this information, the algorithm generates a path from the local position to the destination, minimizing travel costs and the impact of ocean currents.

The resulting path involves waypoints, represented as vector value functions, where the USV navigates while adapting to changing environmental conditions. As seen in Figures 6, the solid path adapts to the changes in ocean currents in both scenarios A and B, effectively minimizing consumption costs. The proportion of consumption is represented in Fig. 6 by the intensity of color in each point of the trajectory, ranging from dark blue to yellow, and is normalized between 0 and 1.

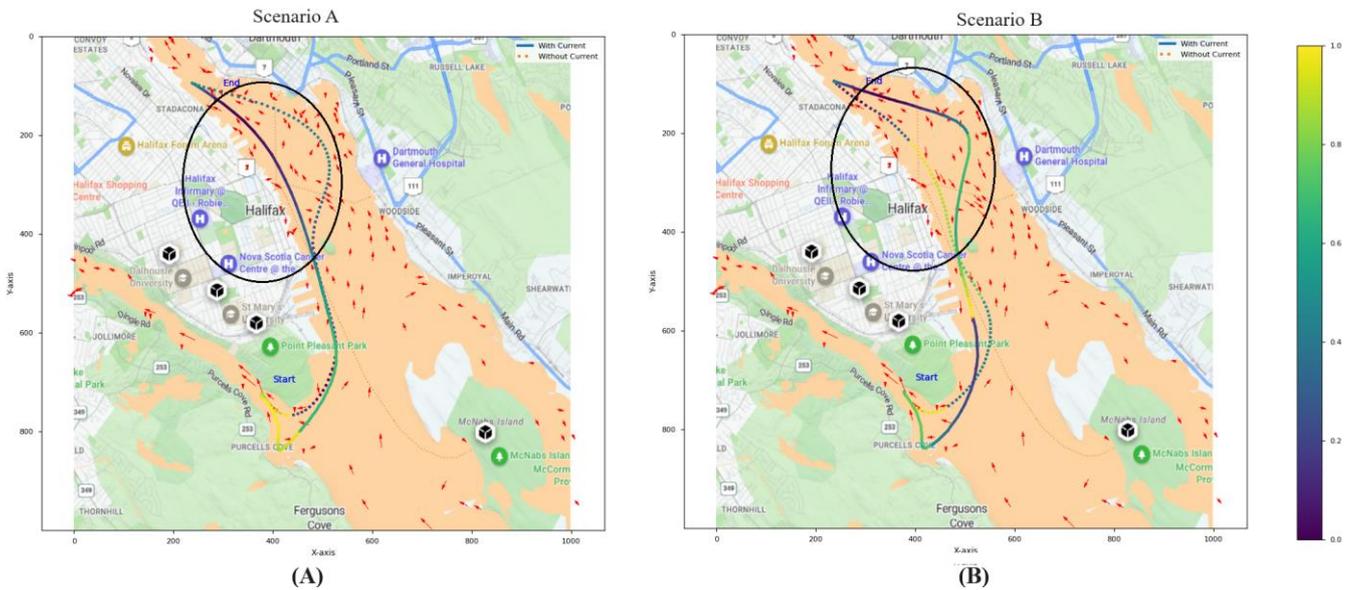

**Fig. 6.** Approximately optimized paths, considering ocean current settings A and B, were analyzed both with (solid curve) and without (dotted curve) the influence of ocean currents. The proportion of the consumption factor is represented by the color intensity. A total decrease of about 7% and 16% in settings A and B, respectively, was observed.

*C. Comparison with AFM*

In this section, we present a comparative analysis of our proposed method and a recently published study known as Anisotropic GPMP2 [6] across a simulated scenario. The Anisotropic GPMP2 approach incorporates AFM to construct a cost function for energy consumption. AFM is a numerical method extensively used in computational geometry and image processing to solve the Eikonal equation while considering the anisotropic characteristics of the environment.

The Eikonal equation is a partial differential equation that describes the propagation of a wavefront through a medium with varying speeds. It calculates the time of arrival (or distance) from a given starting point to all other points in the domain by taking into account the varying speeds associated with each point. AFM, on its own, has the capability to be utilized for path planning, incorporating obstacle avoidance, and minimizing energy consumption. At each location, the local ocean current direction is mathematically represented as an anisotropy, and the energy measurement is determined to assign the highest importance to areas aligned with the ocean current direction. To calculate the cost of the Current in Anisotropic GPMP2, the following steps are needed:

*Step 1*: Make a grid from the environment and assign speeds or costs to each grid cell. Cells with higher costs represent areas where the USV will face more resistance or slower speeds.

*Step 2*: Apply the AFM algorithm to the grid with respect to the source and destination to compute the arrival times from the starting position to all other grid cells and use the result as the cost of the Current.

The utilization of AFM as a cost calculation method in vectorized regions is intriguing. However, the Anisotropic GPMP2 algorithm is susceptible to vulnerabilities based on three primary assumptions. Firstly, AFM assumes that the currents are part of the wavefront, which is not always true, especially in tidal areas. Secondly, in dynamic environments where the Currents constantly change, the algorithm necessitates updating the weights or speeds for each grid cell at every step, introducing potential vulnerabilities. Lastly, when consistent data is unavailable for the entire grid, the algorithm relies on interpolation, leading to computational difficulties, especially in large environments where interpolation must be performed for numerous grid cells. Another vulnerability emerges from the inherent characteristics of AFM when combined with GPMP. AFM's reliance on the potential field of the environment with the destination introduces a potential bias when solely used for estimating current costs within GPMP. The path optimization in GPMP involves iterations to minimize the overall trajectory cost. However, in AFM, there is a possibility that points closer to the destination will consistently have lower costs compared to those closer to the source, which can potentially distort the estimation, Fig. 7. The results show lower precision in areas close to the source and destination, as depicted in Fig. 8.

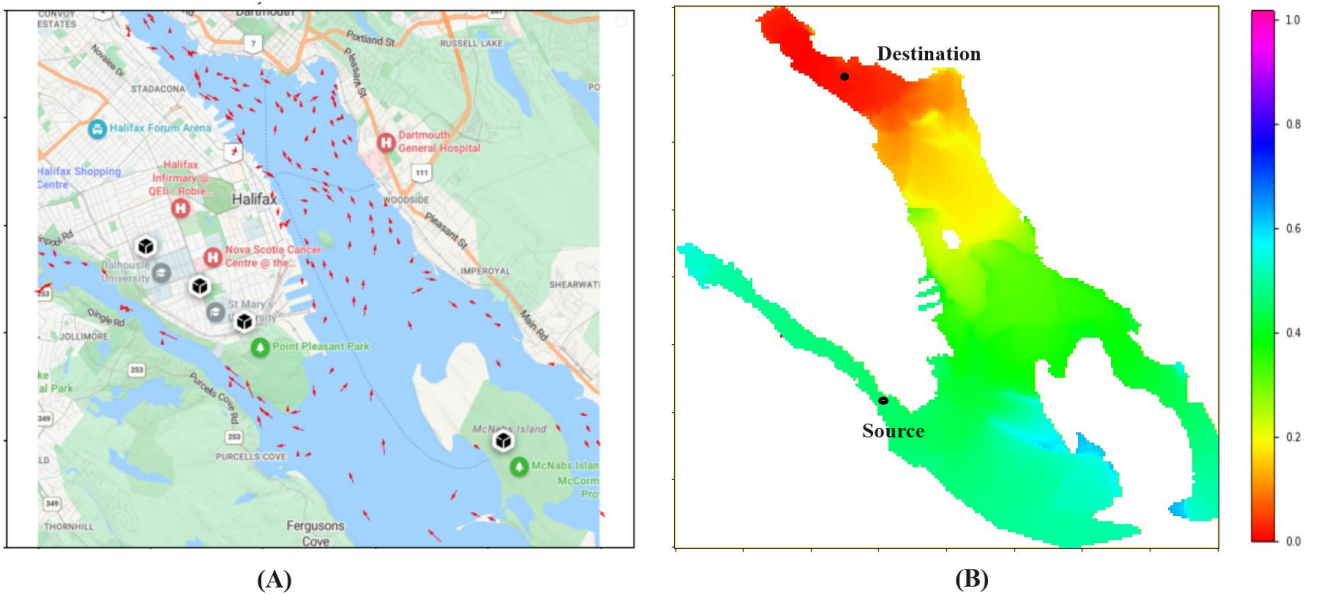

(A)                                      (B)

Fig. 7. Subfigure (A) illustrates the given ocean currents in a certain date; while subfigure (B) displays the AFM arrival time, derived from the linear interpolation of Currents and the destination. The color bar represents the proportion of arrival times as calculated by the AFM.

We conducted a comparison between our algorithm and Anisotropic GPMP2 to demonstrate the impact of bias. The scenario mentioned in Fig. 7 was used for testing. Both algorithms utilized RRT* [13] for generating the initial path. The current information was obtained from the datasets in [26] and [27], and the same arguments were applied to both algorithms. The main difference lies in how trajectory costs were computed. Our algorithm utilized GP inference to track the current, while Anisotropic GPMP2 employed AFM, as shown in Fig. 7(B). In Fig. 8, it's evident that points closer to the source or destination were penalized by their distance to the destination due to the characteristics of AFM. This resulted in some wavering of currents at the beginning and end of the trajectory in GPMP2, compared to our algorithm, which independently computed the cost for each point along the trajectory using GP inference. Due to the non-uniform distribution of points along the two trajectories, we compared each trajectory to the manually selected ground truth by computing the squared loss based on the nearest neighbor. This comparison was facilitated using KDTree in Python. When comparing the residuals to the ground truth path (green path) in Fig. 8(A), our method exhibited an average improvement of 16% in squared loss over 100 Monte Carlo runs.

V. CONCLUSIONS

This study presents a pioneering approach aimed at tracking the vector of energy consumption during movement. It improves upon the GPMP2 model by incorporating a novel spatiotemporal factor, which enables the tracking and prediction of ocean currents using spatiotemporal Gaussian process inference. To validate the effectiveness of the proposed method, experiments were conducted in the demanding environment of Halifax Harbor, specifically targeting the tracking of ocean currents.

As a direction for future research, it would be worthwhile to expand the scope of the model by incorporating additional factors that influence energy consumption, such as water density and wind. By considering these additional factors, the model could offer a more comprehensive understanding of the energy dynamics involved.

The developed methodology holds promise for diverse applications in oceanic studies. Notably, it enables simultaneous tracking of both the path and the currents within the ocean. This opens up possibilities for further investigations and studies in related fields, demonstrating the potential significance of this method in shaping future research endeavors.

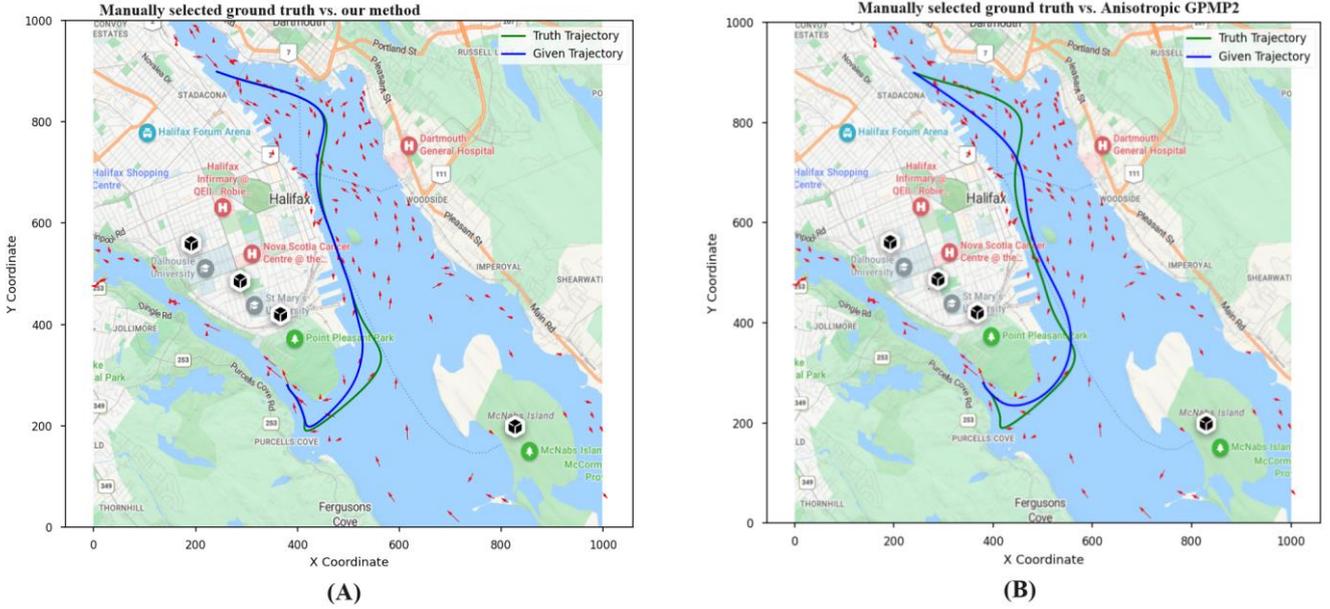

Fig. 8. Path planning for the scenario depicted in Fig. 7. It points out in (B) that in Anisotropic GPMP2, due to a bias, the lower costs assigned to points near the source lead to the USV passing through higher currents.

## APPENDIX A

*1) Ocean Current Inference*

In this section, we explain the process of tracking the ocean current along the trajectory in linear time. A kernel-based GP (Gaussian Process) inference has been used to update the ocean current vector. The majority of GP kernels, including the Current kernels, can be expressed using a straightforward and practical form of the State-Space Differential Equation (SDE), which is linear and time-invariant [24]. By using this model and a recursive Kalman filter, we can estimate the state and covariance in a linear manner at each time step. In order to determine the target association, we utilize a Nearest Neighbor algorithm. The overall recursion for tracking the current vector can be reformulated as Fig.4 and Algorithm 1.

*Temporal Covariance Kernel:* In our application, we used the Matern covariance function for the temporal part. We considered $v = 3/2$, with a continuous and once differentiable process. In this case, the covariance function is simplified to Eq. (25):

$$\kappa_t(\tau)_{\frac{3}{2}} = \sigma^2 \left(1 + \frac{\sqrt{3}\tau}{l}\right) exp\left(-\frac{\sqrt{3}\tau}{l}\right). \quad (25)$$

The continuous system matrix and the noise effect vector of the corresponding state-space model are derived as follows:

$$\mathcal{A}_t = \begin{bmatrix} 0 & 1 \\ -\lambda^2 & -2\lambda \end{bmatrix}, \quad \mathbf{L}_t = \begin{bmatrix} 0 \\ 1 \end{bmatrix}, \quad (26)$$

$$\mathbf{P}_\infty = \mathbf{P}_0 = \begin{bmatrix} \sigma^2 & 0 \\ 0 & \lambda^2 \sigma^2 \end{bmatrix},$$

where $\lambda = \frac{\sqrt{3}\tau}{l}$ and the spectral density of the Gaussian white process noise $w(t)$ is $\mathbf{Q}_c = 4\lambda^3 \sigma^2$. The measurement model matrix is $\mathbf{H} = [1 \ 0]$. The state of a target $i$ can be defined as:

$$\mathbf{v}_i = [[v(ux_1)_k, \dot{v}(x_1)_k], \dots, [v(x_{N^f})_k, \dot{v}(x_{N^f})_k]]. \quad (27)$$

In the discrete case:

$$\mathbf{H}_k = \mathbf{H}_t, \quad \mathbf{L}_k = \mathbf{L}_t, \quad \mathbf{F}_k = e^{\mathcal{A}_t T_s}. \quad (28)$$

Subscript $t$ denotes the temporal continuous model matrices in (20). The spectral density and the initialized state covariance have spatial structures as Eq.(22).

$$\mathbf{Q_k} = (\mathbf{P}_\infty - \mathbf{F}_k \mathbf{P}_\infty \mathbf{F}_k^T), \quad \mathbf{P}_0 = \mathbf{P}_{0,t}. \quad (29)$$

Based on this initial information and using the states and covariance and transition matrices in Eq.(28), we can predict the new state and covariance similar to a multi-variant

Kalman filter. For sudo measurements $\mathbf{z_k(x_k)}$, Eq.(21) is used.

$$\mathbf{v_{k+1|k}} = \mathbf{F_k v_{k|k}} \tag{30}$$
$$\mathbf{P_{k+1|k}} = \mathbf{F_k P_k F_k^\top} + \mathbf{Q_k}, \tag{31}$$
$$\upsilon_k = \mathbf{z_k} - \mathbf{H_k v_{k+1|k}}, \tag{32}$$
$$\mathbf{S_k} = \mathbf{H_k P_{k+1|k} H_k^\top R_k}, \tag{33}$$
$$\mathbf{K_k} = \mathbf{P_{k+1|k} \overline{H}_k^\top S_k^{-1}}, \tag{34}$$
$$\mathbf{v_{k+1}} = \mathbf{v_{k+1|k}} + \mathbf{K_k} \upsilon_k, \tag{35}$$
$$\mathbf{P_{k+1}} = \mathbf{P_{k+1|k}} - \mathbf{K_k S_k K_k^\top}. \tag{36}$$

---

**Algorithm 1.** Step tracking the Current

**Input:**  $\mathbf{Z_{k+1}}, \kappa(x,x'), \mathbf{x_k}, \mathbf{v_k}, \mathbf{P_k}$
**Output:** $\mathbf{v_{k+1}}, \mathbf{P_{k+1}}$ # The Current vector and Covariance
1: $\mathbf{F_k}, \mathbf{H_k}, \mathbf{Q_k}, \mathbf{R} \leftarrow$ Eq. (28) to (29)
2: $\mathbf{v_{k+1|k}}, \mathbf{P_{k+1|k}} \leftarrow$ Eq. (30) and (31)
3: $\mathbf{H}_k^f(\mathbf{x_k}) \leftarrow$ Eq. (24)
4: $\mathbf{R}_k^f(\mathbf{x_k}) \leftarrow$ Eq. (23)
5: $\mathbf{z_{k+1}(x_k)} \leftarrow$ Eq. (21)
6: $\mathbf{v_{k+1}}, \mathbf{P_{k+1}} \leftarrow$ Eq. (33) to (36)
7: **return** $(\mathbf{v_{k+1}}, \mathbf{P_{k+1}})$

---

**Algorithm 2.** Training the Ocean Current Dynamics

**Input:** $\{\mathbf{x}_d^*, \mathbf{z}_d^*\}_{d=1}^D, \kappa(.), \sigma_0^2, l_0, \mathcal{T}$
**Output:** F, L, Qc, H, P_inf, P0
1: $\kappa_t =$ Matérn $(\sigma_0^2, l_0)$  #For the time domain
2: $\kappa_x =$ RBF $(\sigma_0^2, l_0)$   #In the space domain
3: $M_t \leftarrow$ Gpy.GPRegression $(\{\mathbf{x}_d^*, \mathbf{z}_d^*\}_{d=1}^D, \kappa_t)$
4: $M_x \leftarrow$ Gpy.GPRegression $(\{\mathbf{x}_d^*, \mathbf{z}_d^*\}_{d=1}^D, \kappa_x)$
5: $\hat{k}_t \leftarrow M_t$.optimize('BFGS', max_iters)
6: $\hat{k}_x \leftarrow M_x$.optimize('BFGS', max_iters)
7: $l, \sigma \leftarrow \hat{k}_t(l), \hat{k}_t(\sigma)$  #Only for the time domain
8: $\lambda \leftarrow \sqrt{3}\tau/l$
9: $Q_c \leftarrow 4\lambda^3 \sigma^2$
10: $\mathcal{A}_t, L_t, P_\infty \leftarrow$ Eq. (28)
11: $H_k = H_\mathcal{T}, L_k = L_\mathcal{T}, F_k = e^{\mathcal{A}_\mathcal{T} T_s}, P_0 = P_\infty$
12: $Q_k = P_\infty - F_k P_\infty F^\top$
13: **return** $F_k, L_k, Q_k, H, P_0, \hat{k}_x$

ACKNOWLEDGMENT

This work was supported in part by the MITACS Accelerate program of the project named "Intelligent Unmanned Surface Vehicles with Safe Navigation and Docking in Harsh Marine Environment" in collaboration with Marine Thinking Inc, Canada, and the Natural Sciences and Engineering Research Council, Canada.

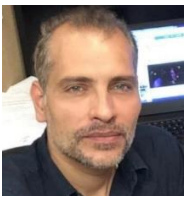
**Behzad Akbari (M'21)**, an IEEE member, earned his bachelor's and master's degrees in Computer Software and Computer Architecture from Tehran and Knowledge And Research University, Iran, in 1996 and 1999, respectively. He further pursued his academic journey by completing a second master's and a PhD in Computer Science and Electrical and Computer Engineering (ECE) from McMaster University, Canada, in 2014 and 2021, respectively. At the onset of this paper, he was a postdoctoral fellow at Dalhousie University, Halifax, Nova Scotia, Canada, in the Department of Mechanical Engineering. He currently serves as an assistant professor at Nipissing University in North Bay, Ontario, Canada. His editorial contributions include IEEE Access, Transactions on Robotics, and Transactions on Systems, Man, and Cybernetics (SMC). Akbari's research is primarily focused on state estimation algorithms, collaborative multi-agent systems, multi-target tracking, multi-output Gaussian processes, iterative localization and mapping, factor graph optimization, and reinforcement learning.

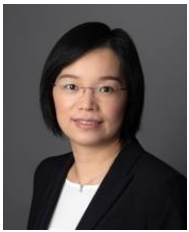
**Ya-Jun Pan (S'00-M'03-SM'11)** is a Professor in the Dept. of Mechanical Engineering at Dalhousie University, Canada. She received the B.E. degree in Mechanical Engineering from Yanshan University, the M.E. degree in Mechanical Engineering from Zhejiang University, and the Ph.D. degree in Electrical and Computer Engineering from the National University of Singapore. She held post-doctoral positions of CNRS in the Laboratoire d'Automatique de Grenoble in France and the Dept. of Electrical and Computer Engineering at the University of Alberta in Canada respectively. Her research interests are robust nonlinear control, cyber physical systems, intelligent transportation systems, haptics, and collaborative multiple robotic systems. She has served as Senior Editor and Technical Editor for IEEE/ASME Trans. on Mechatronics, Associate Editor for IEEE Trans. on Cybernetics, IEEE Transactions on Industrial Informatics, IEEE Industrial Electronics Magazine, and IEEE Trans. on Industrial Electronics. She is a Fellow of Canadian Academy of Engineering (CAE), Engineering Institute of Canada (EIC), ASME, CSME, and a registered Professional Engineer in Nova Scotia, Canada.

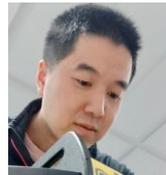
**Shiwei Liu** is Marine Thinking's CTO, who is experienced in developing products for the marine sector with AI and machine learning technologies. He holds a master's degree in computer science from the University of Utah and has over 12 years of experience in Machine Learning, Artificial Intelligence, and Software Engineering fields. Before joining Marine Thinking in 2019, he was a Senior Data Scientist at Vivint Smart Home and led a team of data scientists and engineers to deliver an innovative world-class home security AI solution to clients. He has been granted over 10 US patents, and many others internationally, particularly in Canada and Europe. At Marine Thinking, Shiwei critically evaluates potential projects and makes high-level decisions on their technical feasibility. He is also responsible for technical product planning, leading project teams, and communicating with employees, stakeholders, and customers.

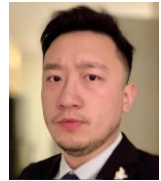
**Tianye Wang** is an AI-Software Lead at Marine Thinking, He received his bachelor's degree and master's degree in computer science from Dalhousie University, Canada. In his role, he leads a team of experts and collaborates with research institutions such as the National Research Council (NRC) and MITACS responsible for developing AI autopilot system for uncrewed surface vehicles (USVs) and machine learning-based solutions on edge devices for the marine sector. His research interests are computer vision, deep learning, edge computing, sensor fusion, autonomous navigation and mapping, swarm robotics, and machine-learning security.